\documentclass[10pt,twocolumn,letterpaper]{article}

\usepackage{iccv}
\usepackage{times}
\usepackage{epsfig}
\usepackage{graphicx}
\usepackage{amsmath}
\usepackage{amssymb}
\usepackage{multirow}

\usepackage[breaklinks=true,bookmarks=false]{hyperref}

\iccvfinalcopy 

\def\iccvPaperID{3036} 

\ificcvfinal\fi

\begin{document}

\title{Vehicle Re-identification with Viewpoint-aware Metric Learning}

\author{Ruihang Chu$^1$, Yifan Sun$^2$, Yadong Li$^3$, Zheng Liu$^3$, Chi Zhang$^{3}$\thanks{Corresponding author} , Yichen Wei$^3$\\
$^1$ School of Mechanical Engineering and Automation, Beihang University\\ $^2$ Department of Electronic Engineering, Tsinghua University\\ $^3$ Megvii Technology\\
{\tt\small {churuihang@buaa.edu.cn sunyf15@mails.tsinghua.edu.cn}}\\
{\tt\small {\{liyadong, liuzheng03, zhangchi, weiyichen\}@megvii.com}}}

\maketitle
\ificcvfinal\thispagestyle{empty}\fi

\begin{abstract}
   This paper considers vehicle re-identification (re-ID) problem. The extreme viewpoint variation (up to 180 degrees) poses great challenges for existing approaches. Inspired by the behavior in human's recognition process, we propose a novel viewpoint-aware metric learning approach. It learns two metrics for similar viewpoints and different viewpoints in two feature spaces, respectively, giving rise to viewpoint-aware network (VANet). During training, two types of constraints are applied jointly. During inference, viewpoint is firstly estimated and the corresponding metric is used. Experimental results confirm that VANet significantly improves re-ID accuracy, especially when the pair is observed from different viewpoints. Our method establishes the new state-of-the-art on two benchmarks.~\footnote{R. Chu and Y. Sun share equal contribution.}\footnote{Work done at Megvii Technology.}
\end{abstract}

\section{Introduction}

\label{sec:intro}

\begin{figure}[t]
	\begin{center}
		\includegraphics[width=1\linewidth,height=0.47\textheight]{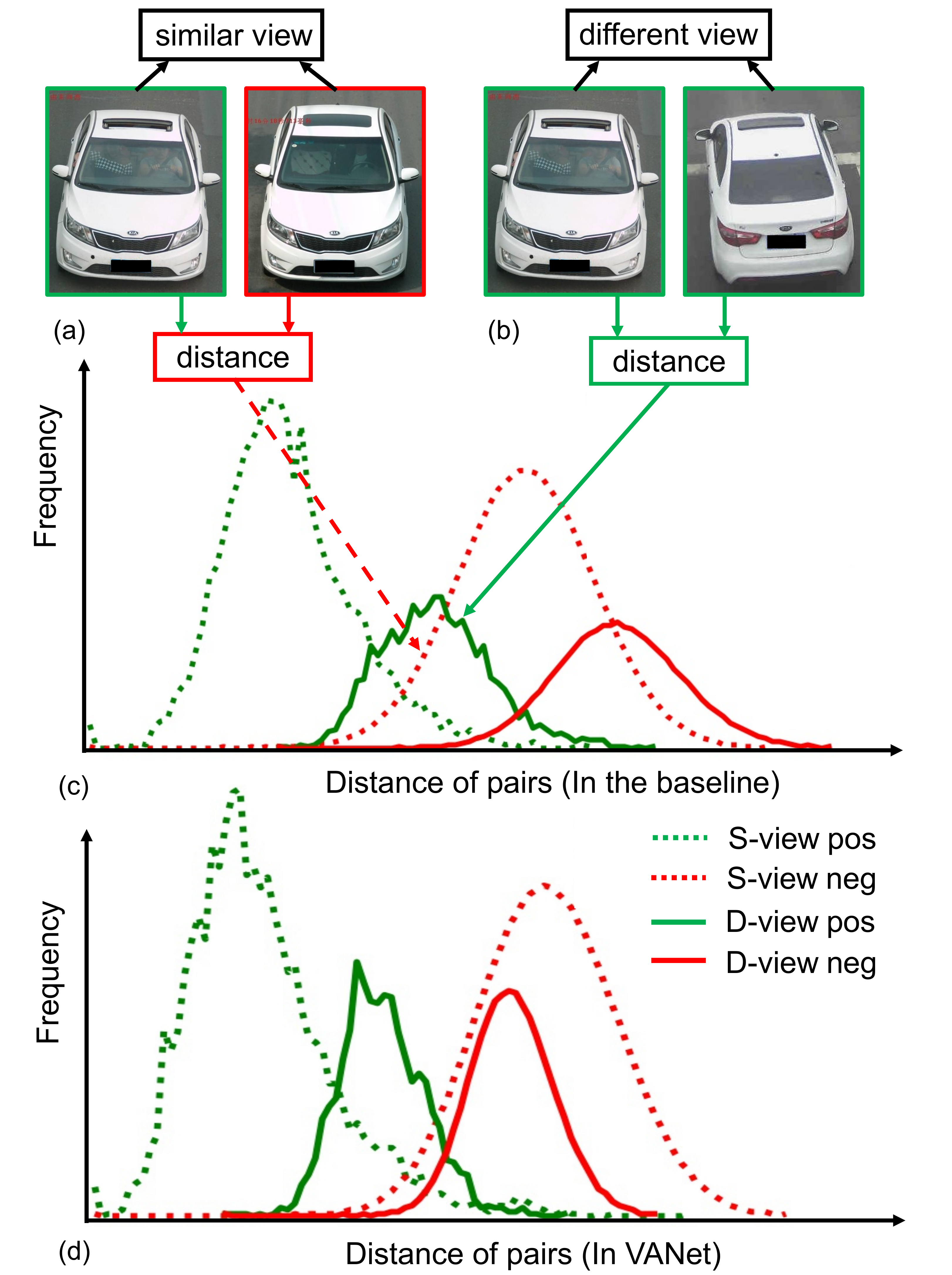}
	\end{center}
	\caption{Illustration of vehicle re-ID with viewpoint variations. Two different vehicles from similar viewpoints (a) appear more similar than the same vehicle observed from different viewpoints (b). (c) distance distribution of a baseline method (see Tab.~\ref{tab:component}). (d) distance distribution of proposed VANet method. S-view pos: positive pairs from S-views. S-view neg: negative pairs from S-views. D-view pos: positive pairs from D-views. D-view neg: negative pairs from D-views. In the baseline (c), the ``D-view pos'' are not well separated from ``S-view neg''. In VANet (d), the distances of ``D-view pos'' become smaller, while the distances of ``S-view neg'' become larger. They are better separated.}
	\label{fig:introduction}
	\vspace{-10pt}
\end{figure}

Vehicle re-identification (re-ID) aims at matching vehicles in surveillance cameras with different viewpoints. It is of significant value in public security and intelligent transportation. The major challenge is the viewpoint variation problem. As illustrated in Fig.~\ref{fig:introduction}, two different vehicles may appear similar from the similar viewpoint (Fig.~\ref{fig:introduction} (a)), while the same vehicle appears quite different from different viewpoints (Fig.~\ref{fig:introduction} (b)). For simplicity, we define \textbf{S-view} as similar viewpoint and \textbf{D-view} as different viewpoint.

Handling viewpoint variations in object recognition has been well studied, \emph{e.g.}, for person re-identification~\cite{Zhao_2017_ICCV,zhou2018graph,Su_2017_ICCV,saquib2018pose,cho2016improving,ma2017pose,Ma_2018_CVPR,liu2018pose,Qian_2018_ECCV} and face verification~\cite{Yim_2015_CVPR,zhu2014multi,Taigman_2014_CVPR,masi2016pose}. While deep metric learning has achieved certain success in obtaining viewpoint invariant features, the viewpoint variation for vehicles is extreme (180 degrees) and this is still very challenging. 

We experimented with a deep metric learning baseline (as detailed in Sec.~\ref{subsec:baseline} and~\ref{sec:dataset}). The distribution of distance of positive pairs from D-views (D-view pos) and distance of negative pairs from S-views (S-view neg) are plotted in Fig.~\ref{fig:introduction} (c). Statistically, many positive pairs with different views have larger distances than those of negative pairs with the same view. This severely deteriorates the retrieval accuracy.

We propose to tackle this challenge through learning viewpoint-aware metrics. Our idea is inspired by mimicking the human behavior of recognizing vehicles. When a human compares two vehicle images, he/she only examines the detailed visual appearance when the vehicles are from S-view. If they are from D-view, he/she relies on association and memorization, instead of directly comparing the visual appearance. In another word, a human adopts different strategies when confronted with vehicle images from S-view and D-view.

We embed this a mechanism in a deep metric learning network, named Viewpoint-Aware Network (VANet) in this work. VANet has two viewpoint-specific metric learning branches, which correspond to two feature spaces where S-view and D-view metrics are learnt, respectively.

Specifically, we enforce two types of metric constraints during training, \emph{i.e.}, within-space constraint and cross-space constraint. The former pushes positive pairs closer than negative pairs in each feature space (corresponding to S-view pos \emph{vs} S-view neg and D-view pos \emph{vs} D-view neg). The latter pushes positive pairs closer than negative pairs when they are in different feature spaces (corresponding to D-view pos \emph{vs} S-view neg and S-view pos \emph{vs} D-view neg). Experiments confirm that both constraints are crucial to the discriminative ability of VANet. They jointly improve vehicle re-ID accuracy significantly, as shown in Fig.~\ref{fig:introduction} (d).

To summarize, our contribution are as follows:

\begin{itemize}
	\item We propose a novel viewpoint-aware metric learning approach for vehicle re-ID. It learns two viewpoint-aware metrics corresponding to S-view pairs and D-view pairs, respectively.
	
	\item We propose two types of constraints, \emph{i.e.}, the within-space constraint and the cross-space constraint for training. The cross-space constraint facilitates retrieving D-view samples under S-view distractions and is especially beneficial. The within-space constraint further brings incremental improvement to our method.
	
	\item {We comprehensively evaluate our method on two large-scale vehicle re-ID datasets and validate the effectiveness of proposed VANet. Experimental results confirm that VANet significantly improves re-ID accuracy over the baseline, and as well outperforms the state of the arts.}
\end{itemize}

\section{Related Works}
\textbf{Vehicle re-ID.} Due to the wide application in urban surveillance and intelligent transportation, vehicle re-ID gains rapidly increasing attention in the past two years. To enhance re-ID capacity, some approaches utilize extra attribute information \emph{e.g.}, model/type, color to guide vision-based representation learning \cite{cormier2016low,liu2016large,liu2018provid,wang2017orientation,shen2017learning,zhou2017cross,Zhou_2018_CVPR,zhou2018vehicle,liu2018ram}. For instance, Liu \emph{et al.} \cite{liu2016ccl} introduce a two-branch retrieval pipeline to extract both model and instance differences. Yan \emph{et al.} \cite{Yan_2017_ICCV} explore multi-grain relationships of vehicles with multi-level attributes. Other works investigate spatial-temporal association, which gains extra benefit from the topology information of cameras \cite{wang2017orientation,shen2017learning,liu2016deep}. In comparison with the above-mentioned works, our method only relies on ID supervision and auxiliary viewpoint information, which is relatively resource-efficient, and yet achieves competitive re-ID accuracy. 

We further compare the proposed VANet with some other works \emph{w.r.t.} viewpoint variation issue. Some works inject orientation information into feature embedding explicitly \cite{sochor2016boxcars} or implicitly \cite{wang2017orientation}. Sochor \emph{et al.} \cite{sochor2016boxcars} embed the orientation vector into feature map and gain viewpoint awareness. Wang \emph{et al.} \cite{wang2017orientation} learn a feature for each side of the vehicle and then credit the feature of visible side during matching. Moreover, some methods employ GAN \cite{zhou2017cross,zhou2018vehicle,Zhou_2018_CVPR} to generate images from required viewpoints, so as to achieve viewpoint alignment. Arguably, these works solve viewpoint variation problem through viewpoint alignment. In contrast, VANet does \emph{NOT} align viewpoints. From another perspective, VANet divides the re-ID into S-view comparison, D-view comparison and learns a respective deep metric for each case. Experimental results confirm the effectiveness of this strategy, especially when confronted with the difficult problem of distinguishing D-view positive samples from S-view negative samples. 

\textbf{Deep metric learning.} Deep metric learning is a common approach employed in computer vision tasks \emph{e.g.}, image retrieval \cite{opitz2017bier, yuan2017hard}, person and vehicle re-identification \cite{bai2018group, cho2016improving, ding2015deep} and face recognition \cite{Schroff_2015_CVPR}. Generally, deep metric learning aims to learn a feature space, in which the samples of a same class are close to each other and the samples of different classes are far away. There are two fundamental types of loss functions for deep metric learning, \emph{i.e.}, the contrastive loss \cite{hadsell2006dimensionality} and the triplet loss \cite{Schroff_2015_CVPR}. The triplet loss is validated as superior to the contrastive loss by a lot of works related to face recognition \cite{Schroff_2015_CVPR,hermans2017defense} and person re-identification \cite{ding2015deep,Cheng_2016_CVPR}. We basis our work on triplet loss function. 

The proposed VANet is featured for two convolutional branches, which correspond to S-view metric and D-view metric, respectively. Experimentally, we demonstrate that viewpoint-aware metrics improve the vehicle re-ID accuracy over the baseline with a single metric. We also confirm that the improvement is largely due to the capacity of VANet to distinguish D-view positive samples from S-view negative samples, as to be detailed in Sec.~\ref{subsec:how}.

\section{Methods}
\label{sec:method}
To learn a respective deep metric for both S-view pairs and D-view pairs, we design a two-branch network to project a single input image into two feature spaces. The distances of S-view pairs and D-view pairs are then measured (through Euclidean distance) in the corresponding feature space, respectively. Sec.~\ref{subsec:baseline} briefly introduces the metric learning baseline. Sec.~\ref{subsec:method} introduces the viewpoint-aware metric learning objective and the corresponding loss function. Sec.~\ref{subsec:arc} implements the viewpoint-aware metric learning with a deep neural network, \emph{i.e.}, VANet.

\subsection{Metric Learning Baseline}
\label{subsec:baseline}

We adopt the commonly-used triplet loss to build the metric learning baseline. Let $\mathcal{X}$ denote the set of data, and $P = (x_i, x_j)$ denote a image pair, where $x_i$ and $x_j\in \mathcal{X}$. Let function $f$ denote the mapping from the raw image to the feature, and $D$ denote the Euclidean distance between features. Given a image pair $P = (x_i, x_j)$, we calculate their distance by $D(P)=D(x_i,x_j)= ||f(x_i)-f(x_j)||_2$.

Following these definitions, given three samples $x$, $x^+$, and $x^-$, where $x$ and $x^+$ are from a same class (\ie, same vehicle ID in our work) and $x^-$ are from another class, we form a positive pair $P^+ = (x, x^+)$ and a negative pair $P^- = (x, x^-)$. A triplet loss is then defined as follows:
\begin{equation}\small
L_{Tri.}(x, x^+, x^-) = \max \{ D(P^+) - D(P^-) + \alpha , 0 \},
\label{equ:triplet}
\end{equation}
where $\alpha$ is a margin enforced between positive and negative pairs. Generally, Eq.~\eqref{equ:triplet} pulls the samples from a same ID close to each other and pushes samples from different IDs far away. However, as shown in Fig.~\ref{fig:introduction} (c), due to the extreme viewpoint variation problem, the baseline fails to separate a large proportion of D-view positive samples from S-view negative samples.

\subsection{Viewpoint-aware Metric Learning}
\label{subsec:method}
From the observation in Fig.~\ref{fig:introduction} (c), we argue that a unique similarity metric may be insufficient for vehicle re-ID in presence of both S-view and D-view pairs. Instead of learning a unique feature for both cases, we propose a viewpoint-specific framework, which learns two separate deep metrics for samples from S-views and D-views, respectively.

Formally, we define two functions $f_s$ and $f_d$, which map the input image into two respective feature spaces, \emph{i.e.}, S-view and D-view feature space, respectively. Consequentially, we calculate the pair-wise distance in S-view and D-view feature space by $D_s(P)=||f_s(x_i)-f_s(x_j)||_2$ and $D_d(P)=||f_d(x_i)-f_d(x_j)||_2$, respectively. We further define $P^+_s$ (S-view positive pair), $P^+_d$ (D-view positive pair), $P^-_s$ (S-view negative pair) and $P^-_d$ (D-view negative pair). 

We find two types of constraints necessary for viewpoint-aware metric learning, \emph{i.e.}, within-space constraints and cross-space constraints. The within-space constraints expect that in both feature spaces, $D(P^+)$ is smaller than $D(P^-)$. The cross-space constraints expect that in each feature space, $D(P^+)$ is smaller than $D(P^-)$ in the other feature space. 
\begin{figure}
	\begin{center}
		\includegraphics[width=1\linewidth]{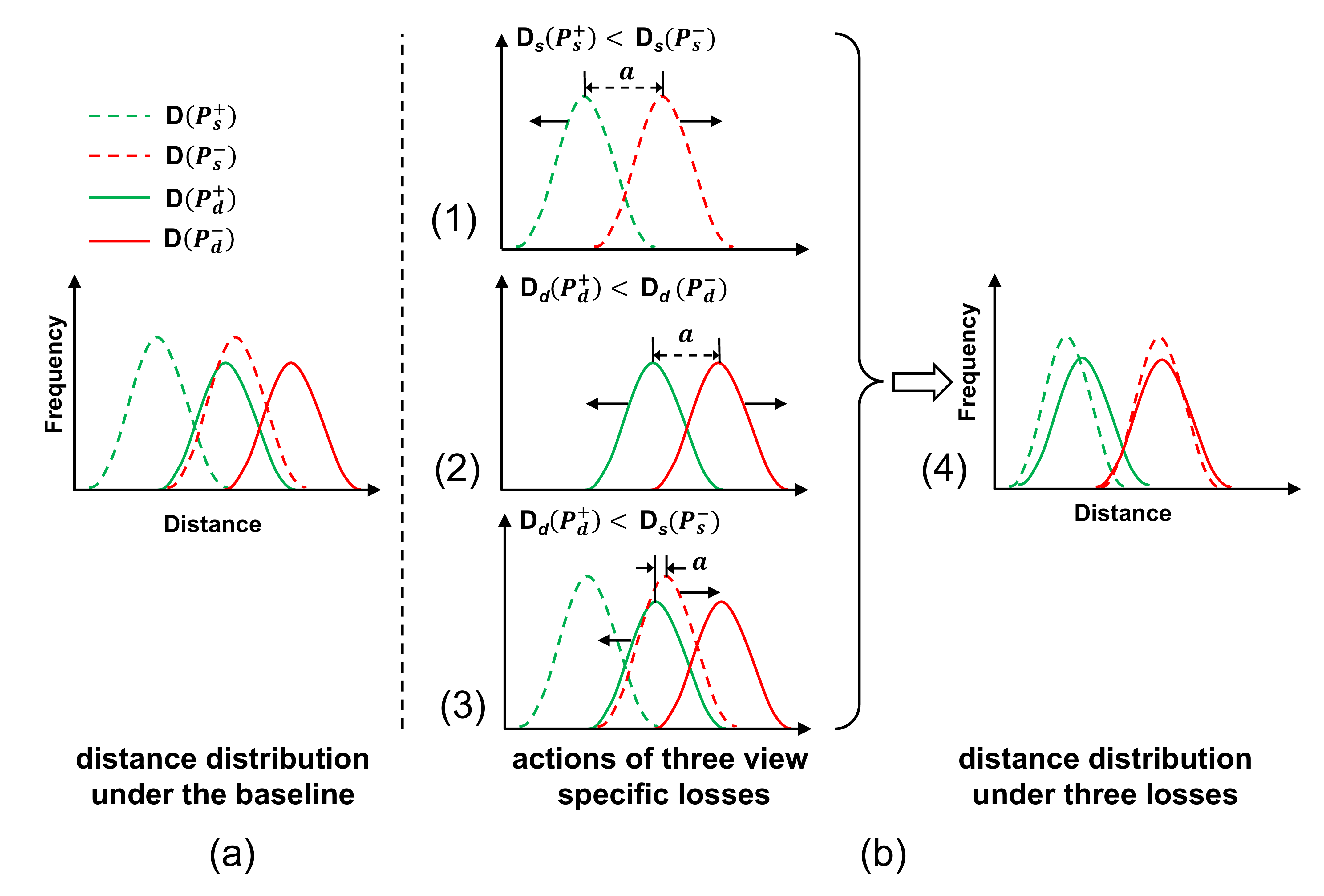}
	\end{center}
	\vspace{-4mm}
	\caption{Comparison of the metric learning in the baseline and our method. (a): in the baseline, the D-view positive samples are hard to be separated from S-view negative samples. (b): three loss functions aim to separate (1) S-view pos and S-view neg (2) D-view pos and D-view neg and (3) D-view pos and S-view neg. The expected distance distributions are in (1)(2)(3) of (b), respectively. They jointly allows accurate blending viewpoint re-ID, and the overall distance distribution is expected to be as in (4).}
	\label{fig:distribution}
	\vspace{-3mm}
\end{figure}

\begin{figure*}
	\begin{center}
		\includegraphics[width=0.95\linewidth]{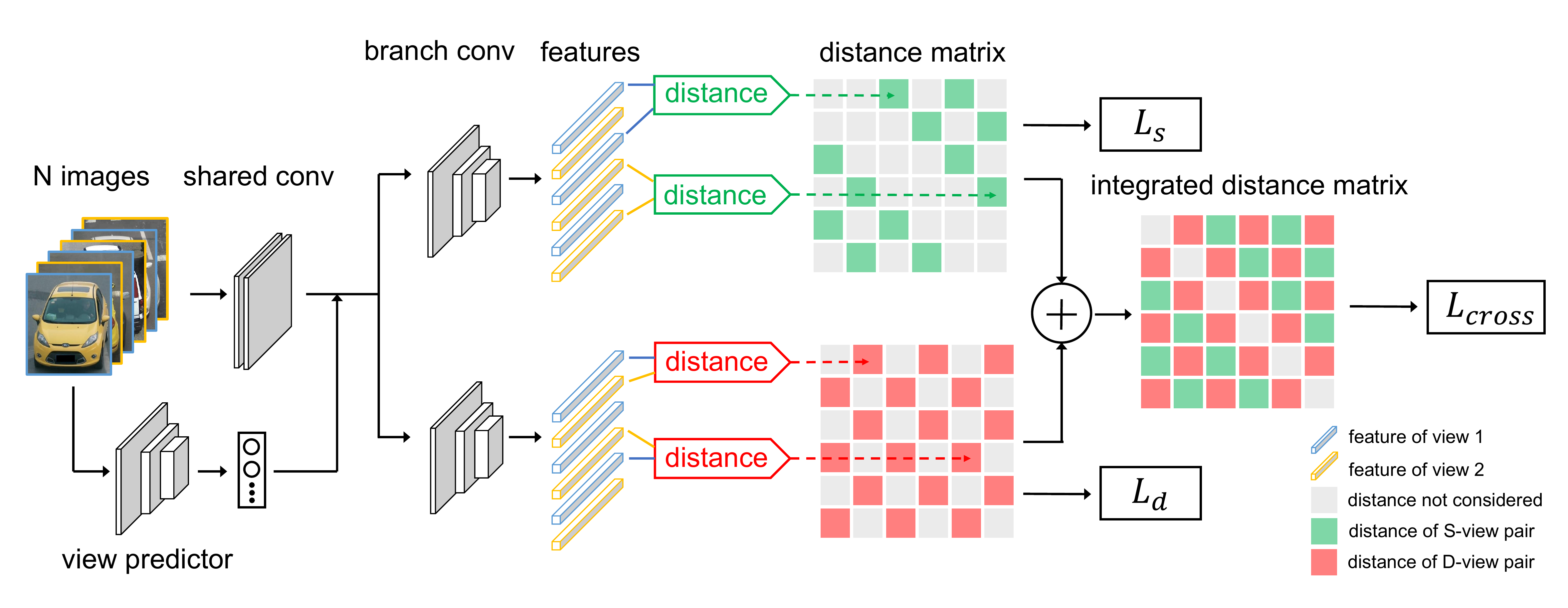}
	\end{center}
   \vspace{-4mm}
	\caption{The architecture of VANet. Firstly VANet employs a viewpoint predictor to predict the actual viewpoints of $N$ images. Then VANet inputs $N$ images into a stack of convolutional layers (``shared conv''), and appends two convolutional branches (``branch conv'') to transform all images to $2N$ features, \emph{i.e.,} $N$ features  in each feature space. Thus VANet generates a $N\times N$ distance matrix in each feature space. In S-view feature space (corresponding to the upper branch), VANet only uses the distances of S-view pairs (green cells) to compute the loss function $L_s$ (defined in Sec.~\ref{subsec:method}). The distances of D-view pairs are omitted. In D-view feature space (corresponding to the bottom branch), VANet only selects the distances of D-view pairs (red cells) to compute $L_d$. Moreover, VANet generates an integrated distance matrix by combining two distance matrices, in which the distances (green and red cells) are used to compute $L_{cross}$.}
	\label{fig:network}
	\vspace{-2mm}
\end{figure*}

\noindent\textbf{Within-space constraints.} It is straightforward that in each feature space, the positive samples should be closer than negative samples. To this end, we enforce the within-space constraints through two triplet losses, one in S-view feature space and one in D-view feature space. The triplet loss in S-view feature space is defined as:
\begin{equation}
L_{s} = \max \{ D_s(P^+_s) - D_s(P^-_s) + \alpha, 0 \},
\label{equ:close}
\end{equation}
and the triplet loss in D-view feature space is defined as:
\begin{equation}
L_{d} = \max \{ D_d(P^+_d) - D_d(P^-_d) + \alpha , 0 \}.
\label{equ:disjoint}
\end{equation}

In each viewpoint-specific feature space, the corresponding loss function only acts on viewpoint-specific samples. In another word, $L_s$ focuses on S-view pairs, while $L_d$ focuses on D-view pairs. In this way, our model can learn to extract discriminative fine-grained visual cues for distinguishing hard vehicles from S-views in the S-view feature space. Meanwhile, our model is able to learn a robust correlation between samples of the same vehicle from D-views in the D-view feature space. 

\noindent\textbf{Cross-space constraints.} Within-branch constraint is not sufficient, since the realistic re-ID system is confronted with a mixture of both cases, \emph{i.e.}, S-view pairs and D-view pairs. Under this settings, focusing within each feature space actually underestimates the viewpoint variation problem and consequentially decreases re-ID accuracy, as to be accessed in Sec.~\ref{subsec:component}. We further propose the cross-space constraints between two viewpoint-specific feature spaces, which are implemented with the following triplet loss:
\begin{equation}
L_{cross} = \max \{ D_d(P^+_d) - D_s(P^-_s) + \alpha , 0 \}.
\label{equ:cross}
\end{equation}

We note that $L_{cross}$ only enforces a constraint between $P_s^-$ and $P_d^+$ and neglects the constraint between $P_d^-$ and $P_s^+$. This is reasonable because two observations of a vehicle from the S-view are prone to be closer than two observations of two different vehicles from D-views.

\textbf{Joint optimization.} 
The total loss triplet function in our work is as follows:
\begin{equation}
L = L_{s} + L_{d} +  L_{cross}
\label{equ:loss}
\end{equation}

The individual action and joint effect of three losses are illustrated in Fig.~\ref{fig:distribution}. By the joint optimization of three losses, the model not only learns from two relatively simple scenarios (\emph{i.e.}, S-view positive \emph{vs} S-view negative and D-view positive vs D-view negative), but also a relatively confusing scenario (\emph{i.e.}, D-view pos \emph{vs} S-view neg).

\subsection{Network Architecture}
\label{subsec:arc}
We implement the metric learning formulated in Sec.~\ref{subsec:method} through VANet, which contains two feature learning branches, as illustrated in Fig.~\ref{fig:network}.


VANet requires to recognize the viewpoint relation (\emph{i.e.}, S-view or D-view) between the input image pair. To this end, VANet first employs a viewpoint classifier to predict the viewpoint of each image, \emph{e.g.}, ``front'' and ``rear''. If two images are captured from the same/similar viewpoint, we assign them as a “S-view” image pair. Otherwise, we assign them as a “D-view” image pair. Specifically, the classifier is trained with the cross-entropy loss using GoogLeNet \cite{szegedy2015going} as the backbone. We manually annotate the viewpoints of 5,000 images from VehicleID \cite{liu2016ccl} and Veri-776 \cite{liu2016deep} respectively as training samples.
The viewpoint labels are available~\footnote{ \url{http://github.com/RyanRuihanG/VANet\_labels}}.
Please note that the viewpoint predictor is trained separately. So the viewpoint predictions are not updated during re-ID feature learning or feature inference.

Then VANet inputs the images into a stack of convolutional layers named ``shared conv". After that, VANet appends two convolutional branches, which are of the identical structure but do not share any parameters. Each branch layer can be regarded as a function for viewpoint-specific feature extraction, \emph{i.e.} $f_s$ and $f_d$ defined in Sec.~\ref{subsec:method}. 
For each image, VANet outputs dual features in difference feature spaces, \emph{i.e.,} both $f_s(x)$ in S-view feature space and $f_d(x)$ in D-view feature space. During training, given a mini-batch consisted of $N$ input images, VANet generates a S-view and a D-view distance matrix in parallel. Each distance matrix contains $N \times N $ elements of distance values. It means that for a D-view image pair, VANet still predicts the S-view distance, \emph{i.e.}, $D_s(P_d)$. Similarly, for a S-view image pair, VANet still predicts the D-view distance, \emph{i.e.}, $D_d(P_s)$. In Fig. \ref{fig:network}, we denote these distance values with grey cells in the distance matrix. We recall that Eq. \eqref{equ:close} (Eq. \eqref{equ:disjoint}) focuses on S-view (D-view) samples. So during training, $D_s(P_s)$ and $D_d(P_d)$ (denoted with the green and red cells in the distance matrices in Fig.~\ref{fig:network}) respectively contribute to the loss $L_s$ and $L_d$, while $D_s(P_d)$ and $D_d(P_s)$ are omitted.

Moreover, VANet selects all $D_s(P_s)$ values from the S-view distance matrix, and $D_d(P_d)$ values from the D-view distance matrix, to form an integrated distance matrix. As shown in Fig.~\ref{fig:network}, in the integrated distance matrix, the green (red) cells corresponds to green (red) cells in S-view (D-view) distance matrix. VANet then further computes triplet loss $L_{cross}$ as Eq.~\eqref{equ:cross} in the integrated distance matrix, enforcing the cross-space constraints.

During testing, given a query image, VANet performs viewpoint-specific comparison with gallery images. Specifically, if the query and the gallery image are predicted as from the S-view, we calculate their distance by $D_s(P_s)$ through S-view branch, otherwise by $D_d(P_d)$ through D-view branch.
\section{Experiments}
\subsection{Datasets and Settings}
\label{sec:dataset}
We conduct extensive experiments on two public benchmarks for vehicle re-ID, \ie, VehicleID \cite{liu2016ccl} and Veri-776 \cite{liu2016deep}. \textbf{VehicleID} is a large-scale vehicle dataset, containing 110,178 images of 13,134 vehicles for training and 111,585 images of 13,113 other vehicles for testing. The test set is further divided into three subsets with different sizes (\ie, \textit{Small}, \textit{Medium}, \textit{Large}).
\textbf{Veri-776} dataset consists of more than 50,000 images of 776 vehicles captured by 20 surveillance cameras, split to a training set containing 37,778 images of 576 vehicles and a testing set including 11,579 images of 200 vehicles.

\textbf{Evaluation protocols.} During evaluation, we follow the protocol proposed in \cite{liu2016ccl,liu2016deep}. On VehicleID, one of the images of a vehicle is randomly selected as the gallery sample and the rest is regarded as query. On Veri-776, we follow the original protocol to retrieve queries in an image-to-track fashion, where queries and the correct gallery samples must be captured from different cameras. We compute the Cumulative Matching Characteristic (CMC) curves for both datasets, and further compute the Mean Precision (mAP) for Veri-776 (which has multiple true matches for a single query). The experiments on VehicleID are repeated 10 times for an average result.

\textbf{Implementation Details.}
We adopt GoogLeNet \cite{szegedy2015going} as the backbone model in our experiments. Specifically, we adopt the all layers before Inception (4a) module as the ``shared conv'', and layers from Inception (4a) model to the global average pooling layer as the ``branch conv''. The images are resized to 224$\times$224 then augmented with color jittering and horizontal flip. We train the model for 200 epoch applying the Adam optimizer \cite{kingma2014adam} with default settings ($\epsilon=10^{-3}, \beta_1=0.9, \beta_2=0,999$). The learning rate is initialized to 0.001 and decreased by a factor of 0.1 after the 80$^{th}$ and 160$^{th}$ epoch. The margin $\alpha$ is set to 0.5. Each batch contains 128 samples (32 IDs, 4 images for each) on VehicleID, and 256 samples (32 IDs, 8 images for each) on Veri-776. During training, we adopt batch hard mining strategy \cite{hermans2017defense} for deducing the triplet loss. 

In addition to the triplet loss, we adopt a cross-entropy loss, following several recent re-ID methods \cite{bai2018group, zhang2019densely}. Specifically, we append an ID-classifier upon the feature-embedding layer. The classifier is implemented with a fully-connected layer and a sequential softmax layer. The softmax output is supervised by the ID label of the training images through the cross-entropy loss. Employing extra cross-entropy loss slightly improves the re-ID accuracy of both VANet and the baseline, which is consistent with \cite{bai2018group, zhang2019densely}.

For viewpoint classifier, vehicles are captured from either front or rear viewpoint in VehicleID, so we define (front-front) and (rear-rear) as the S-view pairs, and (front-rear) and (rear-front) as D-view pairs. As for Veri-776, all images are coarsely categorized into three viewpoints as front, side and rear, then we define (front-front), (rear-rear), and (side-side) as S-view pairs, and (front-side), (front-rear) and (rear-side) as D-view pairs.

\subsection{Evaluation and Ablation Study}
\label{subsec:component}

\begin{table}[tbp]
	\begin{center}
		\renewcommand\arraystretch{1.1}
		\setlength{\tabcolsep}{1.0mm}{
			\begin{tabular}{l|cc|ccc}
				\hline
				\multirow{2}{*}{Method}  & \multicolumn{2}{c|}{VehicleID} &\multicolumn{3}{c}{Veri-776} \\
				\cline{2-6}
				& top1 & top5 & top1 & top5 & mAP  \\
				\hline
				(a) Baseline & 75.23 & 91.84 &  84.68 & 93.74 & 58.75 \\
				(b) VANet w/o $L_{cross}$ & 70.23 & 79.26 & 66.20 & 84.56 & 43.32 \\
				(c) VANet w/o $L_{within}$  & 81.20 & 94.08 & 87.42 & 95.81 & 63.22\\
				(d) VANet
				& \textbf{83.26}& \textbf{95.97}&  \textbf{89.78} & \textbf{95.99} & \textbf{66.34} \\
				\hline 
			\end{tabular}
		}
	\end{center}
\vspace{-1mm}
\caption{Results evaluated on the small test set of VehicleID and Veri-776. We make comparison between VANet, the baseline, VANet w/o $L_{within}$, VANet w/o $L_{cross}$.}
\label{tab:component}
\vspace{-2mm}
\end{table}

We evaluate the effectiveness of VANet by comparing it against the baseline (Sec.~\ref{subsec:baseline}), which adopts the same training settings as VANet. Moreover, to validate the importance of the within-space constraints and the cross-space constraints, we conduct an ablation study by removing the $L_{cross}$ and the $L_{within}$ (\emph{i.e.}, $L_s$ and $L_d$), respectively. The results are summarized in Tab.~\ref{tab:component}, from which we make three observations.

\textbf{First, VANet significantly improves vehicle re-ID  performance over the baseline.} Comparing VANet with the baseline, we observe that VANet gains +8.03\% top1 accuracy on VehicleID and +7.59\% mAP on Veri-776, respectively. It demonstrates that VANet can not only improve the retrieval accuracy at rank-1, but also enhance the capacity in spotting more challenge samples. 

\textbf{Second, the cross-space constraints is vital for re-ID.} Comparing ``VANet w/o $L_{cross}$'' with VANet, we observe a dramatic performance degradation (-13.0\% top1 accuracy on VehicleID and -23.02\% mAP on Veri-776). The performance even drops below the baseline (-5.0\% top1 accuracy). It is reasonable because without cross-space constraints, VANet only learns from two relatively easy scenarios: both the positive pairs and the negative pairs are synchronously observed from S-views (or D-views). We thus infer that cross-space constraints plays an important critical role in enhancing the retrieval capacity.

\textbf{The within-space constraints brings  incremental improvement.} 
Comparing VANet with ``VANet w/o $L_{within}$'', we observe an additional improvement (+2.06\% top1 accuracy on VehicleID and +3.12\% mAP on Veri-776). Specifically, while employing single ``$L_{cross}$'' already brings +5.97\% top1 accuracy improvement on VehicleID over the baseline, adding ``$L_{within}$'' further enhances the top1 accuracy by +2.06\%. It indicates that additional within-space loss further benefits the metric learning in each feature space, thus allows more accurate re-ID. 

\subsection{How VANet Improves Re-ID Accuracy.}
\label{subsec:how}
\begin{table}[tbp]
	\begin{center}
		\renewcommand\arraystretch{1.1}
		\setlength{\tabcolsep}{1.9mm}{
			\begin{tabular}{l|cc|cc}
				\hline
				Method &  top1$_s$ & top1$_d$ & top1$_s^*$ & top1$_d^*$   \\
				\hline
				(a) Baseline &92.23 & 33.53 & 94.01 & 56.09\\
				(b) VANet w/o $L_{cross}$ & \textbf{95.83} & 13.59 & \textbf{96.23} & 67.66 \\
				(c) VANet w/o $L_{within}$& 91.34 & 52.81 & 95.12 & 60.67\\
				(d) VANet & 92.80& \textbf{59.85} & 96.06 &\textbf{67.67} \\
				\hline 
			\end{tabular}
		}
	\end{center}
	\vspace{-1mm}
	\caption{Results evaluated by proposed protocol on the small test set of VehicleID. We make comparison between VANet, the baseline, VANet w/o $L_{within}$ and VANet w/o $L_{cross}$.}
	\label{tab:protocol}
	\vspace{-2mm}
\end{table}

We offer an insight about how VANet improves re-ID accuracy by separating S-view and D-view re-ID scenarios. For simplicity, we conduct the experiments on VehicleID, which has only one single true match for each query. 
we investigate the following two questions:
\begin{itemize}
	\item \textbf{Q1:} What is the performance when the true match is observed from S-view and D-view, respectively. 
	\item \textbf{Q2:} Given the prior that the true match is observed from S-view (D-view), what is the performance when the gallery images are all observed from S-view (D-view).
\end{itemize}

\textbf{Viewpoint-related evaluation protocol.} We propose two specific protocol for \textbf{Q1} and \textbf{Q2}, respectively. 

For \textbf{Q1}, we define top$1_s$ and top$1_d$ by: The top$1_s$ is the matching accuracy at rank-1 when the query and the ground truth are observed from S-views. The top$1_d$ is is the matching accuracy at rank-1 when the query and the ground truth are from D-views. 

For \textbf{Q2}, we define top$1_{s}^*$ and top$1_{d}^*$ by: Given the prior that the true match is observed from S-view (as the query), we remove all the D-view gallery images and calculate the rank-1 matching accuracy to get top$1_{s}^*$. Similarly, given the prior that the true match is observed from D-view, we remove all the S-view gallery images and calculate the rank-1 matching accuracy to get  top$1_{d}^*$. We note that top$1_{d}^*$ can be viewed as top$1_{d}$ under a relatively easier condition, because the distraction from S-view gallery images are eliminated. Similar interpretation can be made upon top$1_{s}^*$ and top$1_{s}$.

With the proposed viewpoint-specific protocol, we conduct another ablation study on VehicleID. The results are summarized in Tab.~\ref{tab:protocol}, from which we draw three observations:

\textbf{First, for the baseline, negative pairs from S-view appear similar and thus lead to false match.} From ``(a) baseline'', we clearly observe a significant gap between top1$_s$ and top1$_d$ (92.23\% \emph{vs} 33.53\%). It indicates that D-view ground truth is much more difficult to retrieve. However, when removing all the S-view gallery images, the matching accuracy significantly increases (from 33.53\% to 56.09\%). It indicates that S-view negative pairs introduce significant distractions to re-ID. 

\textbf{Second, VANet effectively suppresses the S-view distractions, especially when retrieving the D-view true match.}
Comparing ``(d) VANet'' with ``(a) Baseline'', we observe a significant improvement, especially on top1$_d$ (+26.32\%). It indicates that VANet presents much higher discriminative ability when retrieving D-view true match (when exposed to the S-view distractions). Even if we eliminate the S-view distractions (\emph{i.e.}, the top1$_d^*$ case), VANet still exhibits considerable improvement over the baseline (+11.58\%). In contrast, the improvement on top1$_s$ and top1$_s^*$ are relatively minor. We thus conclude that the superiority of VANet (over baseline) mainly originates from its strong capacity to retrieve D-view true match. 

\textbf{Third, the cross-space constraint is the major reason for gaining resistance against S-view distractions.}
Comparing ``(b) VANet w/o L$_{cross}$'' with ``(a) Baseline'', we find that without cross-space constraints, VANet does not increase the top1$_d$, but dramatically decreases it. Recall that top1$_d$ indicates the capacity of retrieving D-view ground truth under S-view distractions. We thus infer that the cross-space constraint is critical for VANet to gain resistance against S-view distractions. It also explains the observation in Tab.~\ref{tab:component} that VANet w/o $L_{cross}$ drops below the baseline.

With the above observations, we confirm that a significant drawback of the baseline is: it lacks the capacity to retrieve D-view ground truth, especially when exposed to S-view distractions. In contrast, the proposed viewpoint-aware metric in VANet significantly improves the discriminative ability to retrieve D-view images, while maintaining (if not slightly improving) the performance under other cases. As a result, VANet significantly improves the overall performance.

\subsection{Visualization of Distance Distribution}
We investigate the learned feature space of VANet and the baseline. Specifically, we calculate the sample distances in the feature space and draw a histogram, as shown in Fig.~\ref{fig:introduction} (c) (the baseline) and (d) (VANet), respectively. From Fig.~\ref{fig:introduction}(c), we observe that in the baseline, the distance ranges of D-view positive pairs and S-view negative pairs are highly overlapped. It implies the baseline cannot separate D-view positive samples from S-view negative samples well in its feature space. In Fig.~\ref{fig:introduction}(d), D-view positive pairs are squeezed to a smaller range (smaller D-view sample distances), and thus are distinctive from S-view negative samples. We thus conclude that VANet presents higher resistance against extreme viewpoint variations.

\begin{figure}[t]
\vspace{-6mm}
	\begin{center}
		\includegraphics[width=1\linewidth]{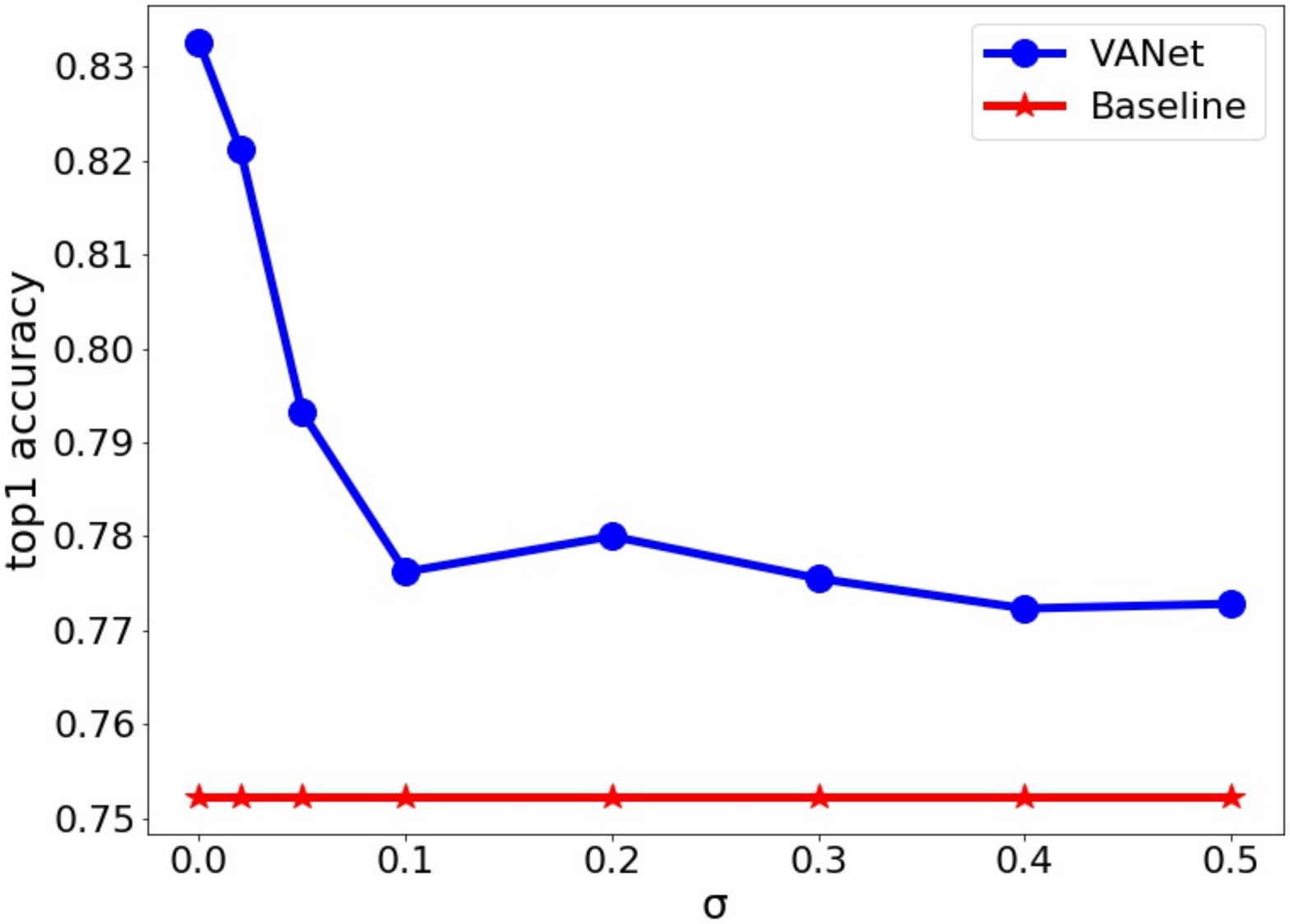}
	\end{center}
\vspace{-5mm}
	\caption{The top1 accuracy evaluated on VehicleID using different viewpoint predictors. The red line denotes the results in the baseline and the blue line denotes the results in VANet. The $\sigma$ denotes the error rate of viewpoint prediction.}
	\label{fig:disturb}
\vspace{-3mm}
\end{figure}

\subsection{Investigations of Viewpoint Prediction}
Viewpoint predictor is a prerequisite component of VANet and its prediction accuracy is important to re-ID accuracy. To validate this point, we train a set of viewpoint predictors with various accuracy by adding error samples. The experimental results are illustrated in Fig.~\ref{fig:disturb}. We observe that higher viewpoint accuracy brings higher re-ID accuracy for VANet. When the error rate $\sigma$ increases to 0.1, the top1 accuracy drops from 83.26\% to 77.63\% (- 5.63\%). It implies that training a viewpoint predictor is important.

In our implementation, we build two different editions of viewpoint predictors with GoogLeNet and Xception \cite{Chollet_2017_CVPR} as the backbone, respectively. 
The GoogLeNet-based predictor achieves 99.4\% accuracy among 2000 randomly selected images on VehicleID and 98.9\% accuracy on Veri-776. The Xception-based predictor only contains 10\% parameters compared with GoogLeNet and achieves roughly the same viewpoint prediction accuracy. We recommend using the Xception-based predictor with consideration of computational efficiency.

\subsection{Impact of Viewpoint Granularity}
\label{subsec:branches}
\begin{table}[tbp]
	\centering
	\renewcommand\arraystretch{1.1}
	\setlength{\tabcolsep}{1.6mm}{
		\begin{tabular}{c|cc|cc|cc}
			\hline
			\multirow{2}*{Branches}  &\multicolumn{2}{c|}{Small} &\multicolumn{2}{c|}{Medium} & \multicolumn{2}{c}{Large}\\
			\cline{2-7}
			& top1 & top5 & top1 & top5 & top1 & top5\\
			\hline
			2&\textbf{83.26}&\textbf{95.97}&81.11&\textbf{94.71}&\textbf{77.21}& \textbf{92.92}\\
			3&83.01&95.80&\textbf{81.23}&94.14&76.89&92.87\\
			4&82.39&94.79&81.11&94.01&76.34&92.08 \\
			\hline
		\end{tabular}
	}
	\vspace{1mm}
	\caption{Results evaluated on the small test set of VehicleID using different branch numbers.}
	\label{tab:num_VehicleID}
	\vspace{-2mm}
\end{table}

In VANet, we only divide the viewpoint relation into two types, \emph{i.e.}, S-view and D-view. To investigate the impact of viewpoint granularity on re-ID accuracy, we introduce several finer-grained divisions to viewpoint relations. On VehicleID, we further construct a three-branch and a four-branch network. Specifically, the thee-branch network learns three different metrics, \emph{i.e.}, (front-front), (rear-rear) and (front/rear) viewpoint metrics. The four-branch network learns (front-front), (rear-rear), (front-rear) and (rear-front) viewpoint metrics explicitly. The experimental results are summarized in Tab.~\ref{tab:num_VehicleID}. It is observed that finer-grained viewpoint division actually compromises re-ID accuracy. Moreover, we implement a six-branch network on Veri-776, which learns (front-front), (rear-rear), (side-side),  (front-rear), (front-side) and (rear-side) metrics. It only achieves 64.35\% mAP, \emph{i.e.}, -1.99\% lower than the two-branch VANet. There are two reasons for this phenomenon. First, since the dataset contains a fixed amount of images, finer division reduces training samples for each branch, exposing each branch to a higher over-fitting risk. Second, as we observe in Fig.~\ref{fig:disturb}, an inaccurate viewpoint predictor compromises the performance of VANet. Because finer division is more likely to increase the viewpoint prediction error, using finer-grained viewpoint division could be worse.

\subsection{Comparison with State of The Art}
\label{subsec:art}

\begin{table*}[!htbp]
	\begin{center}
		\renewcommand\arraystretch{1.1}
		\setlength{\tabcolsep}{3.4mm}{
		\begin{tabular}{l|ccc|ccc|ccc}
			\hline
			\multirow{2}{*}{Methods}  & \multicolumn{3}{c|}{Small} &\multicolumn{3}{c|}{Medium} & \multicolumn{3}{c}{Large}\\
			\cline{2-10}
			&top1 & top5 & top20 & top1 & top5 & top20 & top1 & top5 & top20\\
			\hline
			DRDL \cite{liu2016ccl} &48.93 & 75.65 & 88.47 & 45.05 & 68.85 & 79.88 & 41.05 & 63.38 & 76.62\\
			XVGAN \cite{zhou2017cross}& 52.87 & 80.83 & 91.86 & 49.55 & 71.39 & 81.73 & 44.89 & 66.65 & 78.04\\
			CLVR \cite{kanaci2017vehicle} & 62.00 & 76.00 & - & 56.10 & 71.80 &- & 50.60 &68.00 & - \\
			OIFE$^+$ \cite{wang2017orientation}& - & - & - & - & - & - & 67.00 & 82.90\\
			RAM \cite{liu2018ram} & 75.20 & 91.50 & - & 72.30 & 87.00 & - & 67.70 & 84.50&-\\
			ABLN \cite{zhou2018vehicle}& 52.63 & 80.51 & 91.25 & - & -& - & - & -&-\\
			VAMI \cite{Zhou_2018_CVPR}& 63.12& 83.25 & 92.40 & 52.87 & 75.12 & 83.49 & 47.34 & 70.29&79.95\\
			C2F \cite{guo2018learning}& 61.10& 81.70 & - & 56.20 & 76.20 & - &51.40 & 72.20 & - \\
			NuFACT \cite{liu2018provid}& 48.90 & 69.51 & - & 43.64 & 65.34 & - & 38.63 & 60.72 & -\\
			GSTE(GoogLeNet) \cite{bai2018group} & 77.20 & - & - & 76.40 & - & - & 74.60 & - & -\\
			GSTE(ResNet50) \cite{bai2018group} & 87.10 & - & - & 82.10 & - & - & 79.80 & - & -\\
			\hline 
			\textbf{Baseline} &  75.23 &91.84 &97.01 &73.39 & 87.95&94.84 &69.16 & 85.15&93.48\\
			\textbf{VANet} & 83.26 &95.97 &98.70 &81.11 &94.71 & 98.47&77.21 &92.92 &97.66\\
			\hline
			\textbf{Baseline(ResNet50)} &  78.41 &92.41 &97.59 &75.55 & 91.18 &95.04 &72.38 & 86.74&94.88\\
			\textbf{VANet(ResNet50)} & \textbf{88.12} &\textbf{97.29} &\textbf{99.13} &\textbf{83.17} & \textbf{95.14} & \textbf{98.78} &\textbf{80.35} &\textbf{92.97} &\textbf{98.21}\\
			\hline

		\end{tabular}
	}
	\end{center}
\vspace{-1mm}
\caption{Comparison with state of the art method on VehicleID. ``+'' denotes method that utilize external data for training models.}
\label{tab:vehicleidart}
\vspace{-1mm}
\end{table*}

\begin{table}[!htbp]
	\begin{center}
		\renewcommand\arraystretch{1.1}
		\setlength{\tabcolsep}{4.2mm}{
			\begin{tabular}{l|ccc}
				\hline
				Methods &mAP &top1 & top5\\
				\hline
				FACT+STR \cite{liu2016deep} & 27.77 & 61.44 & 78.78 \\
				XVGAN \cite{zhou2017cross}& 24.65 & 60.20 & 77.03 \\
				OIFE$^+$+ST \cite{wang2017orientation}& 51.42 & 68.30 & 89.70 \\
				S+LSTM \cite{shen2017learning}& 58.27 & 83.49 & 90.04 \\
				RAM \cite{liu2018ram}& 61.50 & 88.60 & 94.00  \\
				ABLN \cite{zhou2018vehicle}& 22.91 & 58.14 & 74.41  \\
				VAMI+STR \cite{Zhou_2018_CVPR} & 61.32& 85.92 & 91.84  \\
				GSTE \cite{bai2018group} & 59.40 & - & - \\
				\hline
				\textbf{Baseline}  &58.75 & 84.68& 93.74 \\
				\textbf{VANet} &  \textbf{66.34}& \textbf{89.78}& \textbf{95.99}\\
				\hline

			\end{tabular}
		}
	\end{center}
	\vspace{-1mm}
	\caption{Comparison with state of the art method on Veri-776.}
	\label{tab:veri776art}
	\vspace{-4mm}
\end{table}
			
\noindent \textbf{Performance on VehicleID.}
We compare our method with several state-of-the-art methods on VehicleID, including DRDL \cite{liu2016ccl}, XVGAN \cite{zhou2017cross}, CLVR \cite{kanaci2017vehicle}, OIFE \cite{wang2017orientation}, RAM \cite{liu2018ram}, ABLN \cite{zhou2018vehicle}, VAMI \cite{Zhou_2018_CVPR}, C2F \cite{guo2018learning}, NuFACT \cite{liu2018provid}, and GSTE \cite{bai2018group}. Among these methods, OIFE \cite{wang2017orientation} utilizes external datasets for training models. All the methods except GSTE utilize additional labels (\eg, vehicle types, brands). On the contrary, our method is only supervised by ID labels. Tab.~\ref{tab:vehicleidart} shows the comparison results on VehicleID. Our method significantly outperforms the competing methods, including those involving additional labeling information. Particularly, for fairly comparison with GSTE which also uses the ResNet-50 \cite{he2016deep} as its backbone model, we further report the results of our method using ResNet-50 as well. Specifically, VANet based on ResNet-50 also gains a significant improvement over the baseline, and outperforms GSTE on three test sets of VehicleID.

\noindent{\textbf{Performance on Veri-776.}}
The compared state-of-the-art methods on Veri-776 include FACT+STR \cite{liu2016deep}, XVGAN \cite{zhou2017cross}, OIFE+ST \cite{wang2017orientation}, S+LSTM \cite{shen2017learning}, RAM \cite{liu2018ram}, ABLN \cite{zhou2018vehicle}, VAMI+STR \cite{Zhou_2018_CVPR}, and GSTE \cite{bai2018group}, where the names with ``+ST'' and ``+STR'' mean that the corresponding methods involve the spatio-temporal information. The results shown in Tab.~\ref{tab:veri776art} demonstrate the superiority of our method in the lower ranks of CMC accuracy and mAP. To compare with RAM, which has the second best performance, VANet achieves higher both top1 accuracy and mAP. The great superiority on mAP than RAM demonstrates VANet successfully promotes the ranks of very hard true samples captured from the different viewpoints with the query. This leads to a great improvement of the matching of D-view pairs, which is a common and useful application in real-world scenes. 

\noindent \textbf{Computational Cost.} When we adopt GoogLeNet \cite{szegedy2015going} as the backbone, adding a branch to  the baseline increases 5.1M (+87\%) parameters and 675 MFlops (+44\%) computational cost. However, VANet is still computational efficient compared with other current methods in Tab.~\ref{tab:vehicleidart} and Tab.~\ref{tab:veri776art}. Among them, we found that VANet has the second fewest parameters (10.9M). GSTE \cite{bai2018group} has smallest parameter size (almost 6M), but it has much worse performance compared to VANet (-6.06\% top1 accuracy on VehicleID and -6.94\% mAP on Veri-776.)

\section{Conclusion}
This paper proposes learning viewpoint-aware deep metrics for vehicle re-identification through a two-branch network named VANet. VANet divides vehicle re-ID into two scenarios, \emph{i.e.}, re-ID from similar viewpoints (S-view) and re-ID from different viewpoints (D-view). Correspondingly, VANet learns two respective deep metrics, \emph{i.e.}, the S-view metric and the D-view metric. By enforcing both within-space constraints and cross-space constraints, VANet improves the re-ID accuracy, especially when retrieving D-view images under exposure to S-view distractions. Experimental results confirm that VANet significantly improves re-ID accuracy, especially when the positive pairs are observed from D-view and the negative pairs are observed from S-view (which is particularly difficult for the baseline method). The achieved performance surpasses all state of the arts.  

{\small
\bibliographystyle{ieee_fullname}
\bibliography{egbib}
}

\end{document}